# Efficient Semantic Segmentation on Edge Devices

**Farshad Safavi**
fsafavi1@umbc.edu
University of Maryland,
Baltimore County

**Aryya Gangopadhyay**
gangopad@umbc.edu
University of Maryland,
Baltimore County

**Irfan Ali**
irfana1@umbc.edu
University of Maryland,
Baltimore County

**Venkatesh Dasari**
venkatd1@umbc.edu
University of Maryland,
Baltimore County

**Guanqun Song**
song.2107@osu.edu
The Ohio State University

**Ting Zhu**
zhu.3445@osu.edu
The Ohio State University

**Maryam Rahnemoonfar**
maryam@lehigh.edu
Lehigh University

*Abstract*—Semantic segmentation works on the computer vision algorithm for assigning each pixel of an image into a class. The task of semantic segmentation should be performed with both accuracy and efficiency. Most of the existing deep FCNs yield to heavy computations and these networks are very power hungry, unsuitable for real-time applications on portable devices. This project analyzes current semantic segmentation models to explore the feasibility of applying these models for emergency response during catastrophic events. We compare the performance of real-time semantic segmentation models with non-real-time counterparts constrained by aerial images under oppositional settings. Furthermore, we train several models on the Flood-Net dataset, containing UAV images captured after Hurricane Harvey, and benchmark their execution on special classes such as flooded buildings vs. non-flooded buildings or flooded roads vs. non-flooded roads. In this project, we developed a real-time UNet based model and deployed that network on Jetson AGX Xavier module.

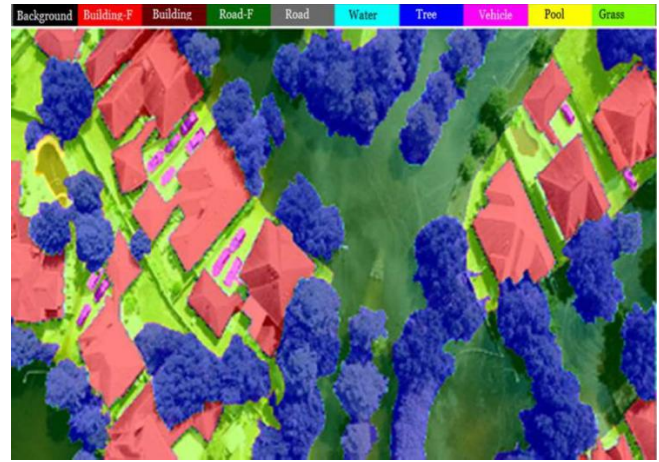

Fig. 1. Semantic segmentation of flooded scene captured by UAV after hurricane Harvey [34],

## I. INTRODUCTION

With the development of science and technology, machine learning techniques have been applied to various fields such as medicine, business, security, IoT and so on [1 – 13], especially in the direction of computer vision and natural language processing have made remarkable achievements [14-20]. Semantic segmentation is a fundamental and challenging task for image understanding. It tries to predict a complicated labeling map for the input image, and assigns each pixel a unique category label. Semantic segmentation has shown great promise in many purposes like autonomous driving and video surveillance. It is essential to understand pictures of aerial imagery for proper emergency reaction during devastating events such as hurricanes, earthquakes, and floods. Unmanned Aerial Vehicles (UAVs) are used to obtain aerial images and scrutinize the context by passing images into a semantic segmentation model for monitoring damaged areas [33].

The main task in semantic segmentation is to divide the aerial scene into de-lineated regions by marking each pixel of an image with a correlated class. This task is essential to spot damaged areas for proper emergency response. As an emergency, semantic segmentation of flooded scene captured by UAV after hurricane might detect a flooded road and recognize flooded buildings from non-flooded building [23].

A modern semantic segmentation constitutes Convolutional Neural Networks (CNNs) to isolate both coarse and fine features of input images. Major architectures extract detailed information and aggregate them with high-level features to obtain higher accuracy [22].

Of late, deep fully convolution network (FCN) based methods have attained remarkable results in semantic segmentation. But existing deep FCNs ache from heavy computations due to a series of high-resolution feature maps for preserving the detailed knowledge in dense estimation. Although lowering the feature map resolution (i.e., applying a large overall stride) via sub-sampling operations (e.g., pooling and convolution striding) can instantly increase the efficiency, it dramatically diminishes the estimation accuracy [24].

However, the up-to-the-minute semantic segmentation models are mainly trained and weighed on ground-based datasets such as Cityscapes, MS-COCO, and Cam-Vid, unsuitable for aerial image segmentations. For example, extracted features from objects in aerial point of view are distinct from objects on the ground view. Hence, neural networks cannot properly segment an aerial scene, especially on deformed or scratched objects during disasters [25].

In this project, we first used Flood-Net datasets to train and evaluate real-time UNet based models for aerial semantic segmentation performances under difficult situations. Then, we use real-time UNet architecture with lightweight encoders, including MobileNetV2 and MobileNetV3, and the decoder and port that on the edge device- NVIDIA Jetson AGX Xavier module to highlight the pros and cons of using these models on such edge devices [26].

Our main contribution is to evaluate the performance of current semantic segmentation models for flooding events using edge devices –NVIDIA Jetson Xavier. In addition, we compare the performance of these models on exceptional classes such as Flooded-Buildings vs. Non- Flooded-Buildings or Flooded-Roads vs. Non-Flooded-Roads [27].

The remainder of this work is organized as follows: Section II describes the motivation and the problem statement of this project. Section III explains the related works, including real-time and non-real-time models and performance of the different architectures on various datasets and edge devices. Section IV describes most popular datasets and Flood Net dataset and explains why it is used as the primary dataset for training semantic segmentation models in our project. Section V describes about NVIDIA Jetson Hardware kit and how it is configured for our project.

Section VI discuss about types of semantic segmentation and their classification. Section VII discuss about non-real time based semantic segmentation and architectures. Section VIII discuss about real time based semantic segmentation and architectures. Section IX describes the detailed structure of different UNET based architectures employed in this project for segmentation tasks. Section X includes the procedure for setting up the NVIDIA Xavier Jetson board to perform semantic segmentation and evaluate its performance. Section XI explains how to install py-torch libraries and dependencies. Section XII provides discussions about our results including performance of accuracy and efficiency along with some of the new findings. Section XIII talks about the issues and challenges we faced during our project. Finally, Section XIV delivers the conclusion and our future work.

## II. MOTIVATION

### A. What's the problem

In this project, we explored the performance of efficient semantic segmentation model on an edge device, especially NVIDIA Jetson AGX Xavier developer kit. We wanted to explore efficient algorithms which are deployable on GPU-based hardware platform and compare their performance on NVIDIA Jetson AGX Xavier developer kit.

### B. Why it's an important problem

With many applications of semantic segmentation like detection of flooded roads and buildings and distinguishing between natural water and flooded water, it can be used for efficient damage assessment tasks. By the advent of neural networks, deep learning methods are used for semantic segmentation tasks, which are power hungry and unsuitable for real-time applications on portable devices. In this project we have benchmarked some of the efficient semantic segmentation models on NVIDIA Jetson AGX Xavier developer kit.

## III. RELATED WORK

In [1], the authors proposed a unique attention based dual supervised decoder for RGBD semantic segmentation. The encoder is designed with a simple but effective attention-based multi-modal fusion module to extract and fuse deeply multi-level paired matching information. They also introduce a dual-branch decoder to effectively influence the correlations and complementary cues of different tasks. Major experiments on NYUD-v2 and SUN-RGBD datasets shows good performance of the architecture compared to modern methods.

The authors designed a simple symmetric and effective network to efficiently use the multilevel cross-modal information for RGBD semantic segmentation. The proposed model has a two-stream encoder and a new dual-branch decoder with the primary designed for semantic segmentation.

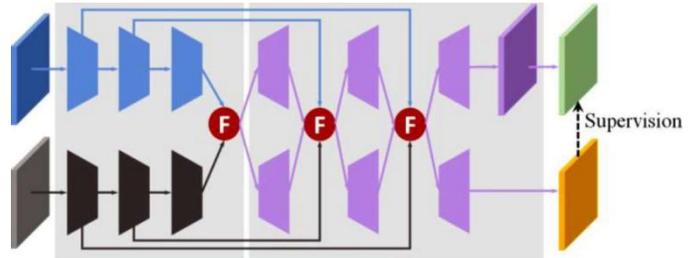

Fig. 2. Dual supervised decoder for segmentation.

In recent years, CNN-based methods have been successfully applied to the R-GBD semantic segmentation. In terms of structure, these methods can be roughly divided into the following three groups [21].

### A. Atrous/dilated Convolution

They can be used to include multi-scale context for R-GBD semantic segmentation or enhance the receptive field while preserving the resolution of the feature map. However, dilated convolution can lose the continuity of feature maps. In addition, it is only efficient for some large objects and unacceptable for small objects, which is not great to extract accurate edges.

### B. Encoder-decoder.

Many encoder-decoder architectures have indulged to RGBD semantic segmentation. Deconv-Net comprise of stacked deconvolutional layers to produce high-resolution prediction and more semantic details. Seg-Net shares a similar scheme using indices in pooling layers to promote the recovery process. RDF-Net elaborates the core idea of residual learning to RGBD semantic segmentation. An architecture AC-Net with three parallel branches and a channel attention-based module abstracts weighted features from R-GB and depth branches. A spatial information guided convolution network (SG-Net) integrates 2D and 3D spatial information. ESA-Net uses two Res-Net-based encoders with an attention-based fusion for indulging depth information, and a decoder comprising a learned-up sampling.

However, these methods only carry out the multi-modal information in the encoder but neglects the cross-modal cues in the decoder. Moreover, when many encoder parameters are passed to the decoder, training such a model becomes difficult to converge quickly.

### C. Multi-task Learning

Numerous works have also explored the idea of joining networks for similar tasks to improve learning efficiency and generalization across different tasks. A single multi-scale network (MSCNN) focus on three different computer vision tasks. A joint task-recursive learning (TRL) framework polished the results of both semantic segmentation and monocular depth estimation using serialized task-level interactions. A pattern affinitive propagation (PAP) method carry out the complemented affinity information across tasks. An intra-task and inter-task pattern-structure diffusion (PSD) is used to realize long-distance propagation and transfer cross-task structures. The authors in [21] integrates multi-modal information in the both encoder and decoder over attention-based dual supervised decoder to provide a unified pixel-wise scene understanding.

In [22], the authors proposed Efficient Boundary-Aware Network (EBA-Net) which count on both RGB and depth images as input. They developed a boundary attention branch for boundary features of objects in the scene and to generate boundary labels for supervision by a Canny edge detector. They also assumed a hybrid loss function blending Cross-Entropy (CE) and structural similarity (SSIM) loss to guide the network to educate the transformation between the input image and the ground truth at the pixel and patch level. They have developed EBA-Net on the common RGB-D dataset NYU-v2 and have reached pretty performance.

In [23], the authors used a dataset of UAV images pointing to different floods taking place in Spain, and then use an AI- based approach that believes on three widely used deep neural networks (DNNs) for semantic segmentation of images, to automatically guess the regions more affected by rains (flooded areas). The targeted algorithms were augmented for GPU-based edge computing platforms, so that the classification can be done on the UAVs themselves, and only the algorithm output is uploaded to the cloud for real-time chasing of the flooded areas. The authors reported advanced real-time processing of UAV images using sophisticated DNN-based solutions.

In [24], the authors studied the extensive research work of semantic segmentation methods based on deep learning reported in the literature and pointed the latest research progress of semantic segmentation outcomes with weakly supervised, domain adjusted, multi-modal data fusion, and real-time.

In [25], the authors present a survey, targeting mainly on the recent scientific developments in semantic segmentation, more on deep learning-based methods using 2-D images. The researchers analyzed the public image sets and leaderboards for 2-D semantic segmentation, with an overall of the techniques employed in performance evaluation. In examining the evolution of the field, they chronologically categorized the techniques into three main periods, namely pre-and early deep learning age, the fully convolutional age, and the post-FCN era.

They scientifically analyzed the solutions put forward in terms of solving the basic problems of the field, such as fine-grained localization and scale invariance. They concluded the survey by discussing the current challenges of the field and to what extent they have been solved.

In [26], the authors target on the deployment of a convolutional neural network for semantic segmentation of urban scenes, to detect hurdles in an outdoor setting. A U-Net-similar architecture was trained on GPU over multiple datasets, typical of the final environment. The selected trained network was then adjusted to perform inference on the Nvidia Jetson Nano hardware accelerator, to be mounted directly on the wheelchair. The outcome achieves an accuracy of around 85% and inference time of 35ms, thus providing a concrete solution towards the target supported power wheelchair.

In [27], the authors deployed three network architectures to predict the road lane: CNN Encoder Decoder network, CNN Encoder Decoder network with the use of Dropout layers and CNN Encoder LSTM-Decoder network that are trained and tested on a public dataset including 12764 road images under variable conditions. Practical outcomes show that the proposed hybrid CNN Encoder–LSTM-Decoder network unified into a Lane-Departure-Warning-System (LDWS) achieves high prediction performance namely an average accuracy of 96.36%, a Recall of 97.54%, and a F1-score of 97.42%. A NVIDIA Jetson Xavier NX super-hardware had been used, for its performance and efficiency, to incorporate an Embedded Deep LDWS.

In [28], the authors assess Recurrent Neural Networks (RNNs), 3D Convolutional Neural Networks (CNNs), and optical flow for video semantic segmentation in parameters of accuracy and inference speed on three datasets with different camera motion arrangements.

The results show that using an RNN with convolutional operators is superior to all methods and achieves a performance boost of 10.8% on the KITTI (MOTS) dataset with 3 degrees of freedom (DoF) motion and a small 0.6% improvement on the Cyber-Zoo dataset with 6 DoF motion over the single-frame-based semantic segmentation method. The inference speed was calculated on the Cyber-Zoo dataset, achieving 321 fps on an NVIDIA GeForce RTX 2060 GPU and 30 fps on an NVIDIA Jetson TX2 mobile computer. The authors show that the best activity relative to the baseline is achieved by leveraging the temporal capability of a simple RNN and LSTM.

In [29], the authors deployed a Multi-Attention-Network (MA-Net) to extract contextual dependencies through multiple efficient attention modules. A new attention mechanism of kernel attention with linear complexity is recommended by the authors to reduce the large computational demand in attention. Based on kernel attention and channel attention, the authors integrate local feature maps extracted by ResNeXt-101 with their corresponding global dependencies and reweight interdependent channel maps adjustably. Several experiments on three large-scale fine resolution remote sensing images captured by different satellite sensors shows the better performance of the proposed MA-Net, outperforming the Deep-Lab V3+, PSP-Net, Fast-FCN, DA-Net, OCR-Net, and other benchmark tactics.

In [30], the authors proposed a design of an emerging network for road extraction based on spatial enhanced and densely connected UNet, called SDU-Net. SDU-Net collects both the multi-level features and global pre information of road networks by combining the strengths of spatial CNN-based segmentation and densely connected blocks. To enhance the feature learning about pre information of road surface, a structure preserving model is designed by the authors to explore the continuous clues in the spatial level. Experimental results on two benchmark datasets shows that the suggested method by the authors achieves the state-of-the-art performance, compared with previous paths for road extraction.

In [31], the authors anticipated a Tensor-RT-based framework supporting various optimization constraints to accelerate a deep learning application targeted on NVIDIA Jetson embedded platform with heterogeneous processors including multi-threading, pipelining, buffer assignment, and network duplication. Since the design space of allocating layers to diverse processing elements and optimizing other parameters is huge, the authors formulate a constraint optimization methodology that consists of a heuristic for harmonizing pipeline stages in heterogeneous processors and fine-tuning process for enhancing constraints. With nine real-life benchmarks, the authors could generate 101% ~ 680% performance upgrade and up to 55% energy reduction over the baseline inference using GPU only.

In [32], the authors reviewed the best technologies of semantic segmentation based on deep learning. Because semantic segmentation requires several pixel-level annotations, to lower the fine-grained requirements of annotation and lessen the economic and time cost of manual annotation, they focus on weakly-supervised semantic segmentation. To improve the generalization ability and robustness of the segmentation model, their research investigates the domain adaptation in semantic segmentation. Several sensors are equipped in some practical applications, such as autonomous driving and medical image analysis.

To associate between multi-modal data and recover the accuracy of the segmentation model, their research investigates on multi-modal data fusion semantic segmentation. The real-time utility of the model needs to be considered in practical application. Their work analyzes the key factors impacting the real-time performance of the segmentation model. Finally, their research enumerates the challenges and promising research directions of semantic segmentation tasks based on deep learning.

In [33], the authors proposed an efficient and robust RGB-D segmentation approach that can be optimized to a high degree using NVIDIA Tensor-RT and, thus, is well adapted as a common starting step in a complex system for scene analysis on portable robots. The authors show that RGB-D segmentation is superior to processing RGB images solely and that it can still be performed in real time if the network architecture is carefully designed. The authors evaluate our proposed Efficient Scene Analysis Network (ESA-Net) on the common indoor datasets NYUv2 and SUNRGB-D and show that the authors reach state-of-the-art performance while enabling faster inference. Furthermore, the authors evaluation on the outdoor dataset Cityscapes shows that our approach is suitable for other areas of application as well. Finally, instead of presenting benchmark results only, the authors also show qualitative results in one of our indoor application scenarios.

In [34], the authors analyze current semantic segmentation models to explore the viability of applying these models for emergency response during catastrophic events like floods. They compared the performance of real-time semantic segmentation models with non-real-time equivalents constrained by aerial images under oppositional settings.

In [35], the authors present a high-resolution UAV imagery, Flood Net, captured after the hurricane Harvey. This dataset demonstrates the after flooded damages of the affected areas. The images are marked pixel-wise for semantic segmentation task and questions are generated for the task of visual problem answering. Flood Net poses several challenges including detection of flooded roads and buildings and distinguishing between natural water and flooded water. With the advancement of deep learning algorithms, the authors can deplore the impact of any disaster which can make a precise understanding of the affected areas. In their research, the authors compare the performances of baseline methods for image classification, semantic segmentation, and visual problem answering on their dataset.

In [36], the authors propose an efficient and effective architecture with a moderate trade-off between speed and accuracy, termed Bilateral Segmentation Network (Bi-Se-Net V2). This architecture indulges:

(i) a Detail Branch, with larger channels and narrow layers to catch low-level details and get high-resolution feature representation.

(ii) a Semantic Branch, with narrow channels and deep layers to get high-level semantic perspective.

The Semantic Branch is lighter due to lower channel capacity and a fast-down sampling strategy. The scientists designed a Guided Aggregation Layer to enlarge mutual connections and join both types of feature representation. Besides, a immunized training strategy is designed to improve the segmentation performance without any extra inducing cost.

In [37], the authors announce a family of efficient backbones specially designed for real-time semantic segmentation. The proposed deep dual-resolution networks (DDR-Nets) constitute two deep branches where multiple bilateral fusions are accomplished. Additionally, the researchers design a new appropriate information extractor dubbed Deep Aggregation Pyramid Pooling Module (DAPPM) to enlarge effective receptive fields and join multi-scale context based on low-resolution feature maps. The researchers were able to trade-off between accuracy and speed on both Cityscapes and Cam-Vid dataset. On a single 2080Ti GPU, DDRNet-23-slim yields 77.4% m-IoU at 102 FPS on Cityscapes test set and 74.7% m-IoU at 230 FPS on Cam-Vid test set. With widely used test augmentation, their methodology is best to most new models and requires much lower computation.

In [38], the authors present Mobile-Nets based on a combination of complementary search techniques as well as a new architecture design. Mobile-NetV3 is tuned to mobile phone CPUs through a combination of hardware aware network architecture search (NAS) supplemented by the Net-Adapt algorithm and then afterwards improved through new architecture designs. This research demonstrates how automated search algorithms and network design can work in friendship to harness complementary methods improving the overall technique. Here the authors create two new Mobile-Net models for release: Mobile-NetV3-Large and Mobile-NetV3-Small which comply towards high and low resource use cases.

These models are then adjusted and applied to the chores of object detection and semantic segmentation. For semantic segmentation (or any dense pixel prediction), the authors offer a new efficient segmentation decoder Lite Reduced Atrous Spatial Pyramid Pooling (LR-ASPP). The authors achieve good results for mobile classification, detection, and segmentation. Mobile-NetV3-Large is 3.2% more accurate on ImageNet classification while lowering latency by 20% compared to Mobile-NetV2. Mobile-NetV3-Small is 6.6% more better compared to a MobileNetV2 model with nearby latency. Mobile-NetV3-Large detection is over 25% faster at nearly same accuracy as Mobile-NetV2 on COCO detection. Mobile-NetV3-Large LRASPP is 34% faster than Mobile-NetV2 R-ASPP at compared accuracy for Cityscapes segmentation.

In [39], the authors propose a new ascending method that uniformly scales all dimensions of depth/width/resolution using a simple and effective composite coefficient. The authors demonstrate the goodness of this method on scaling up Mobile-Nets and Res-Net.

They deploy neural architecture search to implement a new baseline network and scale it up to get a family of models, called Efficient-Nets, which have much better accuracy and efficiency than previous Conv-Nets. Their EfficientNet-B7 has 84.3% top-1 accuracy on ImageNet, while being 8.4x smaller and 6.1x faster on implication than the best existing Conv-Net. Their Efficient-Nets also comply well and achieve best accuracy on CIFAR-100 (91.7%), Flowers (98.8%), and 3 other transfer learning datasets, with an order of magnitude less parameters.

TABLE I. EFFICIENT-NET PERFORMANCE RESULTS ON IMAGE-NET

| Model | Top-1 Acc. | Top-5 Acc. | #Params | Ratio-to-Efficient-Net | #FLOPs |
|---|---|---|---|---|---|
| EfficientNet-B0 | 77.1% | 93.3% | 5.3M | 1X | 0.39B |
| ResNet-50 | 76% | 93% | 26M | 4.9X | 4.1B |
| DenseNet-169 | 76.2% | 93.2% | 14M | 2.6X | 3.5B |
| EfficientNet-B1 | 79.1% | 94.4% | 7.8M | 1X | 0.70B |
| ResNet-152 | 77.8% | 93.8% | 60M | 7.6X | 11B |
| DenseNet-264 | 77.9% | 93.9% | 34M | 4.3X | 6.0B |
| Inception-v3 | 78.8% | 94.4% | 24M | 3X | 5.7B |
| Xception | 79.0% | 94.5% | 23M | 3X | 8.4B |
| EfficientNet-B6 | 84% | 96.8% | 43M | 1X | 19B |

In [40], the authors mention a knowledge distillation method configured for semantic segmentation to improve the performance of the compact FCNs with large overall stride. To handle the contradiction between the features of the student and teacher network, the researchers optimize the matching features in a transferred latent domain formulated by deploying a pretrained autoencoder. Moreover, an affinity distillation module is announced to contain the long-range dependency by calculating the non-local interactions across the whole image.

To substantiate the effectiveness of their suggested method, several experiments have been conducted on three popular benchmarks: Pascal VOC, Cityscapes and Pascal Context. Built upon a highly viable baseline, their proposed method improves the performance of a student network by 2.5% (mIOU boosts from 70.2 to 72.7 on the cityscapes test set) and can train a advance model with only 8% float operations (FLOPS) of a model that achieves good performances.

In [41], the authors states that memory traffic for accessing intermediate feature maps can cause inference latency in real-time object detection and semantic segmentation of high-resolution video. The researchers deploy a Harmonic Densely Connected Network for better efficiency in terms of both low MACs and memory traffic. The new network has 35%, 36%, 30%, 32%, and 45% inference time reduction compared with FC-DenseNet-103, DenseNet-264, ResNet-50, ResNet-152, and SSD-VGG, using tools including Nvidia profiler and ARM Scale-Sim to measure the memory traffic and verify that the inference latency is very proportional to the memory traffic consumption and the proposed network consumes low memory traffic. The researchers conclude that one should take memory traffic into consideration when designing neural network architectures for high-resolution applications at the edge.

In [42], the authors suggest that semantic segmentation needs both rich spatial information and sizeable receptive field. But these techniques compromise spatial resolution to achieve real-time inference speed, causing poor performance. The authors address this issue with a new Bilateral Segmentation Network (BiSeNet). The authors first design a Spatial Path with a small stride to retain the spatial information and generate high-resolution features. Also, a Context Path with a fast downsampling strategy is used to obtain sufficient receptive field. On top of the two paths, new Feature Fusion Module corporates features efficiently. The proposed architecture makes a nice balance between the speed and segmentation performance on Cityscapes, CamVid, and COCO-Stuff datasets. Specifically, for a 2048 X 1024 input, they achieve 68.4% Mean IOU on the Cityscapes test dataset with speed of 105 FPS on one NVIDIA Titan XP card, which is much faster than the existing methods with comparable performance.

In [43], the authors propose a describe a new mobile architecture, MobileNetV2, that improves the state of the art performance of mobile models on multiple tasks and benchmarks as well as across a spectrum of different model sizes. The researchers also describe efficient ways of applying these mobile models to object detection in a novel framework we call SSDLite. Additionally, they demonstrate how to build mobile semantic segmentation models through a reduced form of DeepLabv3 which they call Mobile DeepLabv3. This is based on an inverted residual structure where the shortcut connections are between the thin bottleneck layers.

The intermediate expansion layer uses lightweight depthwise convolutions to filter features as a source of non-linearity. Additionally, the researchers find that it is important to remove non-linearities in the narrow layers in order to maintain representational power. They demonstrate that this improves performance and provide an intuition that led to this design.

Finally, their approach allows decoupling of the input/output domains from the expressiveness of the transformation, which provides a convenient framework for further analysis. They measure the performance on ImageNet classification, COCO object detection, VOC image segmentation. The researchers evaluate the trade-offs between accuracy, and number of operations measured by multiply-adds (MAdd), as well as actual latency, and the number of parameters.

In [44], the authors explained the importance of residual connection which had been studied extensively in their research. Their result shows that the shortcut connecting bottleneck perform better than shortcuts connecting the expanded layers. The linear bottleneck models are strictly less powerful than models with non-linearities, because the activations can always operate in linear regime with appropriate changes to biases and scaling. Their experiments also indicate that linear bottlenecks improve performance, providing support that non-linearity destroys information in low-dimensional space.

In [45], the authors explained a network and training strategy that is based on strong use of data augmentation to use the available annotated samples more efficiently. The architecture consists of a shrinking path to capture the context and a symmetric expanding path that enables precise localization. They show that such a network can be trained end-to-end from very less images and better the prior best method (a sliding-window convolutional network) on the ISBI challenge for segmentation of neuronal structures in electron microscopic stacks. Moreover, the network is fast. Segmentation of a 512x512 image takes less than a second on a new GPU.

## IV. DATASET DESCRIPTION

The level of success for any machine-learning application is undoubtedly determined by the quality and the depth of the data being used for training [1]. For deep learning, data is very important and as most systems are end-to-end; features are determined by the quality of data in the case of deep learning [22].

### A. PASCAL Visual Object Classes (VOC)

This image dataset comprises of image labeling for semantic segmentation, classification, detection, action-classification, and person layout duties. The image set and annotations are regularly updated, and the leaderboard of the challenge is public (with more than 140 submissions just for the segmentation challenge alone).

It is a famous semantic segmentation challenges and is still following its initial release in 2005. The PASCAL VOC semantic segmentation contest image dataset comprises 20 foreground item classes and one background class. The original data contains 1,464 images training, and 1,449 images for validation. The 1,456 test images are hidden for the challenge. The image set includes all types of indoor and outdoor images and is generic across all categories [25].

### B. Common Objects in Context (COCO)

With 200K labelled images, 1.5 million object instances, and 80 object categories, COCO is a very large-scale object detection, semantic segmentation, and captioning image set, including almost every possible type of scene. COCO supports contests at the instance-level and pixel-level stuff-semantic segmentation, and introduces panoptic segmentation, which tries in unifying instance-level and pixel-level segmentation tasks. Their leaderboards have lesser crowding because of the scale of the data [33].

On the other hand, similarly, their trials are assessed only by the most daring scientific and industrial groups, and thus are considered as the best in their leaderboards. Due to its extensive volume, most studies partially use this image set to pre-train or fine-tune their model, before submitting to other challenges such as PASCAL VOC 2012 [25].

TABLE II. ACCURACY AND EFFICIENCY RESULTS FOR SOME REAL-TIME SEMANTIC SEGMENTATION METHODS ON CITYSCAPES TEST DATASET

| Method | Publish | Backbone | Pretrain | Input Size | Params | FLOPs | GPU | FPS | m-IoU (%) |
|---|---|---|---|---|---|---|---|---|---|
| DFANet | CVPR2019 | Xception | ImageNet | 1024 X 1024 | 7.8 M | 3.4G | TitanX | 100 | 71.3 |
| SwiftNet | CVPR2019 | ResNet-18 | ImageNet | 1024 X 2048 | 11.8 M | 104G | 1080Ti | 39.9 | 75.5 |
| ICNet | ECCV2018 | PSPNet-50 | ImageNet | 1024 X 2048 | 26.5 M | 28.3G | TitanX M | 30.3 | 69.5 |
| BiSeNet | ECCV2018 | Xception-39 | ImageNet | 768 X 1536 | 5.8 M | 14.8G | TitanX | 106 | 68.4 |
| BiSeNet | ECCV2018 | ResNet-18 | ImageNet | 768 X 1536 | 12.9 M | 55.3G | TitanX | 65.5 | 74.7 |
| GUN | BMVC2018 | DRN-D-22 | ImageNet | 512 X 1024 | | | TitanX | 33.3 | 70.4 |
| ShelfNet | ICCVW2019 | ResNet-18 | ImageNet | 1024 X 2048 | | | 1080Ti | 36.9 | 74.8 |
| FANet | ECCVW2020 | ResNet-18 | ImageNet | 1024 X 2048 | | 49.0G | TitanX | 72.0 | 74.4 |
| SFNet | ECCVW2020 | ResNet-18 | ImageNet | 1024 X 2048 | 10.5 M | | 1080Ti | 53.0 | 77.8 |

### C. Cityscapes

This is a largescale image set for semantic interpretation of urban lane scenes. It contains annotations for high-resolution images from 50 different cities, taken at different hours of the day and from all seasons of the year, and with varying backgrounds and scene layouts.

The annotations are deployed at two quality levels: refined for 5,000 images and coarse for 20,000 images. There are 30 variable class labels, some of which also have instance labels (vehicles, people, riders etc.). Consequently, there are two challenges with separate public leaderboards: one for pixel-level semantic segmentation, and a second for instance-level semantic segmentation. There are more than 100 entries to the challenge, making it the most popular regarding semantic segmentation of urban street scenes [25].

### D. Scan Net

It is an RGB-D video dataset containing 2.5 million views in more than 1.5k scans. Scan-Net annotated with 3D camera poses, surface reconstructions and instance-level semantic segmentations, collected by an easy-to-use and scalable RGB-D capture system that includes automated surface reconstruction and crowdsourced semantic annotation.

### E. Wood Scape

It is the first autonomous driving fisheye dataset collected by 4 fisheye cameras on the vehicle. Wood-Scape provides semantic annotation of 40 classes at the instance level over 10k images and provides annotations for other tasks for images over 100k [31].

*F. KITTI*

It contains real image data collected from urban, rural and expressway scenes. The diversity of its recording conditions is relatively low only during the daytime and on sunny days in a city. They comprise of 15 vehicles and 30 pedestrians in each image and various degrees of obscuring and truncation [32].

Compared to the KITTI-2012, KITTI-2015 comprises dynamic scenes for which a semi-automatic process establishes the ground truth. KITTI-360 annotates both static and dynamic 3D scene elements with rough bounding primitives and transfers this information into the image domain, resulting in dense semantic and instance annotations for 3D point clouds and 2D images [31].

*G. Inria Aerial Image Labeling Dataset*

It is a remote sensing image dataset for urban building detection. Its labels are allotted for building and non-building, which are mainly deployed for semantic segmentation. The images comprise different urban settlements, ranging from densely populated areas to alpine towns (e.g., Lienz in Austrian Tyrol). Inria Aerial Image Labeling Dataset contains aerial orthophoto corrected color images with a spatial resolution of 0.3 m, and the coverage of Inria Aerial Image Labeling Dataset is 810 km [30].

*H. Flood Net*

Flood Net [35] provides an appropriate benchmark for Realtime semantic segmentation of UAV images. The images are labeled pixel-wise for the semantic segmentation task. Flood Net is a high-resolution aerial imagery dataset for scene understanding, and it is suitable for training real-time semantic segmentation of aerial images. It contains aerial images captured by DJI Mavic Pro quadcopters after Hurricane Harvey. Hurricane Harvey was a devastating flooding event that occurred near Texas and Louisiana in August 2017.

It consists of imagery from UAVs during a disaster, and data reflect the actual situation during flooding events. The aerial images are taken at 200 feet above ground level displaying great details. It contains 2343 high-resolution images and offers semantic segmentation labels for nine different objects, including building-flooded, building-non-flooded, road-flooded, road-non-flooded, water, tree, vehicle, pool, and grass [35].

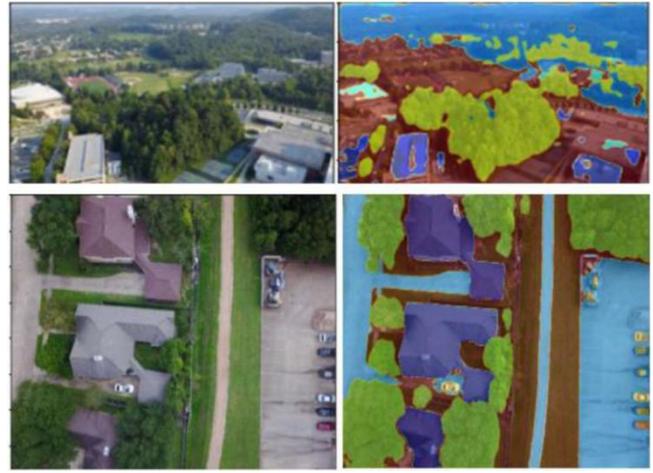

Fig. 3. Real-time semantic segmentation trained on FloodNet dataset is generalized on videos which have different views [33 34].

Publicly available ground imagery datasets such as ImageNet, Microsoft COCO, PASCAL VOC, Cityscapes accelerate the advanced development of current deep neural networks, but the annotation of aerial imagery is scarce and more tedious to obtain [35]. Most of these datasets does not point to the unique contests in understanding after disaster scenarios as a task for disaster damage evaluation.

For fast response and recovery in large scale after a natural calamity such as hurricane, wildfire, and extreme flooding, obtaining aerial images are very important for the reaction team. Flood Net dataset associated with three different computer vision tasks namely classification, semantic segmentation, and visual question answering fills the gap [35].

V. JETSON AGX XAVIER DEVELOPER KIT DESCRIPTION

The NVIDIA Jetson AGX Xavier Developer Kit is the best addition to the Jetson platform. It is an AI computer-hardware for autonomous machines, comparing the performance of a GPU workstation in an embedded module under 30W. Jetson AGX Xavier is devised for robots, drones, and other autonomous machines. With the NVIDIA Jetson AGX Xavier developer kit, researchers can easily deploy end-to-end AI robotics functions for manufacturing, delivery, retail, agriculture, and more. The Nvidia Jetson AGX Xavier (JAX) is one of those accelerators which are built for machine learning with ease of development and deployment of DL applications. Its CPU-GPU architecture supported by CUDA, cu-DNN and the Tenso-RT software libraries for extrapolation speed up makes it the best fit to run DL workloads in several domains such as logistics, facility, manufacturing, smart city deployments and medical image assessment [27].

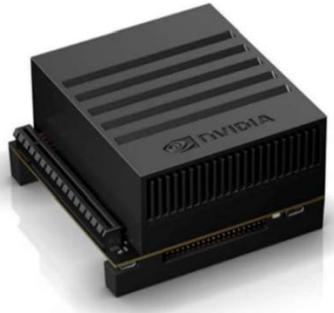

Fig. 4.  NVIDIA Jetson Xavier Module

Backed by NVIDIA Jetpack and Deep Stream SDKs, as well as CUDA®, cu-DNN, and Tensor-RT software libraries, the kit accumulates all the tools researchers need to get started right away. As it is powered by the new NVIDIA Xavier processor, researchers can have more than 20X the performance and 10X the energy efficiency of its predecessor, the NVIDIA Jetson TX2 [27].

The board deploys an 8-core ARMv8 CPU, one Volta GPU, and two NVIDIA DLAs. The DLA is a power-efficient accelerator, but its computational power is less than the Volta GPU. The Xavier has a unified memory shared by CPU and GPU. It means that communication between CPU and GPU can be easily conducted by memory operation but with a potential risk of access contention. When using a DLA, extra communication overhead occurs due to data copy performed internally in Tensor-RT [28].

The board deploys a mode switching alternative according to the power economy for power management. In each power mode, the number of available processors and their frequencies are set differently. Since the performance varies according to the power mode, we need to consider the trade-of between performance and power to optimize both metrics together [31]. Nvidia Jetson is appropriate when power consumption, size, and the ability to perform computer vision tasks in real-time are scarce.

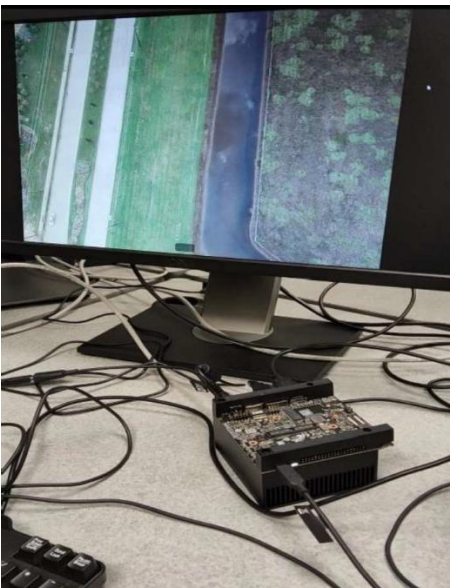

Fig. 5.  NVIDIA Jetson Xavier Module connected to the Host Computer.

## VI. TYPES OF SEMANTIC SEGMENTATION

The purpose of semantic segmentation is to divide an obtained image into many visually meaningful or interesting areas for subsequent image analysis and visual understanding. Semantic segmentation plays an important role in a broad range of applications, e.g., scene understanding, medical image analysis, robot perception and satellite image segmentation. Before applying convolutional neural networks (CNN), researchers used random forest and conditional random field (CRF) to construct classifiers for semantic learning [29].

In contemporary years, the deep learning method has conceded a new generation of segmentation models with good performance improvements and has become the mainstream solution for semantic segmentation. Fully convolutional- network (FCN), deployed end-to-end, when compared with traditional methods, has a 20% improvement in Pascal VOC 2012 dataset. U-net contains a framework path to learn context information and a spatial path to preserve spatial information [32].

TABLE III.  CLASSICAL MODELS OF SEMANTIC SEGMENTATION BASED ON DEEP LEARNING.

| Method | Publish | Year |
|---|---|---|
| FCN | CVPR | 2015 |
| U-net | MICCAI | 2015 |
| Deconv-Net | ICCV | 2015 |
| SegNe | TPAMI | 2015 |
| ERF-Net | TITS | 2018 |
| PSP-Net | CVPR | 2017 |
| Deeplab v1 | ICLR | 2015 |
| Deeplab v2 | TPAMI | 2018 |
| Deeplab v3 | arXiv | 2016 |
| Deeplab v3+ | ECCV | 2018 |
| Dilated Convolutions | ICLR | 2016 |
| RefineNet | CVPR | 2017 |
| DUC | WACV | 2018 |
| ICNet | ECCV | 2018 |
| BiSeNet | ECCV | 2018 |
| CCNet | ICCV | 2019 |
| AdaptSegNet | CVPR | 2018 |
| EncNet | CVPR | 2018 |
| Large Kernel Matters | CVPR | 2017 |

### A. Weakly Supervised Semantic Segmentation.

The fully supervised semantic segmentation method deploying CNNs needs large time and economic cost to obtain pixel-level annotation, which constraints the further progress of segmentation performance and the generalization ability of the model. Therefore, researchers point to weakly-supervised learning, which can take benefit of weak annotation forms and reduce the marking cost [39].

### B. Domain Adaptation Semantic Segmentation.

Most machine learning techniques assume that the training and test set are separate and identically distributed. Obviously, it is difficult to satisfy this notion in practical application. The purpose of the semantic segmentation method based on domain adaptation is to solve the problem of distribution mismatch between training data and test data so that the model can be well extended to practical applications [40].

### C. Semantic Segmentation Based on Multi-Modal Data Fusion.

The advancement in sensors, e.g., cameras, LiDAR, endorses the rapid development of semantic segmentation. Semantic segmentation deploying multi-modal data fusion has potential of new research direction to use data of variety of sensors with opposite characteristics to enhance the performance of segmentation [40].

The sensing system deployed on a single camera cannot provide consistent 3D geometry and adjust to complex or harsh lighting conditions. LiDAR can deliver high-correctness 3D geometry without changing the ambient light, but LiDAR is constraint by low resolution, less refresh rate, severe weather conditions (Rainstorm, fog, and snow), and measurable cost. By blending different types of sensor data, the performance and robustness of semantic segmentation can be enhanced [41].

### D. Real-Time Semantic Segmentation.

The Real-time semantic segmentation is a challenging task as it considers both accuracy and inference speed simultaneously. However, under resource constraints, it is impossible to maintain accuracy and boost inference speed simultaneously. Therefore, real-time semantic segmentation aims to achieve a reasonable trade-off between accuracy and inference speed according to the application's requirements, e.g., autonomous driving [42].

## VII. NON-REAL TIME BASED ARCHITECTURES

Non-real time methods focus on accuracy rather than portability. State-of-the-art non-real time segmentation methods are classified into attention and non-attention-based methods. Non-attention-based methods are further classified into pyramid and encoder-decoder type architecture. Pyramid-based methods implement multi-scale processing of input images to generate feature maps at different spatial scales. On the other hand, methods with encoder-decoder type architecture are composed of an encoder and a decoder. While the encoder performs down sampling to capture semantic or contextual information, the decoder performs up sampling operations to retrieve the spatial data [43].

Attention-based methods became popular after their implementation, and excellent performance in natural language processing. Later its implementation in computer vision has been explored by different researchers and found similar good applications. Attention-based methods attempt to calculate spatial relationships among pixels. Although initially self-attention-based methods were explored for computer vision tasks, including semantic segmentation, recently vision transformers, built upon the self-attention mechanism, are also being applied.

## VIII. REAL TIME BASED ARCHITECTURES

Incoming complex network architecture, the semantic segmentation method based on CNNs suffers from high computational complexity, which greatly limits the application in real-time processing of real scenes, e.g., autonomous driving, video surveillance, robot sensing. In U-oriented structures, e.g., U-net, the encoder pertains to a backbone, which is the most vital part of the whole structure and accounts for the leading proportion of model size and computational budget, the pipeline of U-shape architecture. For real-time inference, some researchers adopt a lightweight backbone model and investigate how to improve the segmentation performance with limited computation. According to the adopted backbone, current real-time semantic segmentation methods can be divided into lightweight classification model-based method, specialized backbone-based method, and two-branch architecture-based method [31].

### A. Bilateral Segmentation Network (Bi-Se-Net)

It has 2 components: Spatial-Path (SP) and Context-Path (CP). These two components are mechanized to tackle with the loss of spatial information and shrinkage of receptive field, respectively. For Spatial Path, researchers pile only three convolution layers to acquire the 1/8 feature map, which retains prosperous spatial details. For Context Path, researchers affix a global average pooling layer on the tail of X-ception, where the receptive field is the highest of the backbone network [36].

In chase of better accuracy without loss of speed, researchers also research the fusion of two paths and refinement of final prediction and propose Feature Fusion Module (FFM) and Attention Refinement Module (ARM) respectively. These two extra components can further improve the whole semantic segmentation accuracy on both Cityscapes, Cam-Vid, and COCO-Stuff benchmarks [36].

### B. BiSeNetV2

This architecture consists of a Detail Branch and a Semantic Branch, which are merged by an Aggregation Layer.

The Detail Branch is liable for the spatial details, which is low-level information. Therefore, this branch entails a rich channel space to encode comfortable spatial detailed information. As the Detail Branch simply concentrates on the low-level details, researchers design a superficial structure with a small stride for this branch. Generally, the key concept of the Detail Branch is to use broad channels and narrow layers for the spatial elements. Besides, the showcase representation in this branch has huge spatial size and broad channels [37]. Therefore, it should not adopt the residual connections, which raises the memory access cost and reduce the speed.

In analogous to the Detail Branch, the Semantic Branch is designed to secure high-level semantics. This branch has minimal channel capacity, while the spatial details can be generated by the Detail Branch. In contrast, in our experiments, the Semantic Branch has a ratio λ (λ < 1). outlets of the Detail Branch, which makes this branch lighter. The Semantic Branch can be any lightweight convolutional model. Meanwhile, the Semantic Branch adopts the fast-down sampling strategy to promote the level of the feature representation and enlarge the receptive field quickly. High-level semantics require large receptive field. Therefore, the Semantic Branch employs the global average pooling to embed the global contextual response [38].

The feature interpretation of the Detail Branch and the Semantic Branch is balancing, one of which is unaware of the information of the other one. Further, an Aggregation Layer is intended to merge both types of feature representation. Due to the quick-down sampling strategy, the spatial dimensions of the Semantic Branch's yield are smaller than the Detail Branch. There is a need to up sample the output feature map of the Semantic Branch to match the output of the Detail Branch. There are less means to fuse information, e.g., simple summation, concatenation, and countable well-designed operations. The BiSeNetV2 framework is a generic architecture, which can be implemented by most convolutional models and achieves a good trade-off between segmentation accuracy and inference speed [39].

### C. Harmonic Densely Connected Network (Hard-Net)

The researchers deploy a new metric for assessing a convolutional neural network by approximating its DRAM traffic for feature maps, which is a vital factor affecting the power consumption of a system. When the intensity of computation is low, the traffic can dictate inference time more significantly than the model size and operation count. They deploy Convolutional Input/Output (CIO) as an rough calculation of the DRAM traffic, and propose a Harmonic Densely Connected Networks (Har-D-Net) that achieve a high accuracy-over-CIO and a high computational efficiency by increasing the density of computation (MACs over CIO). FC-HarD-Net to attain DRAM traffic reduction by 40% and GPU supposition time reduction by 35% contrasted with FC-Dense-Net [44].

Comparing with DenseNet-264 and ResNet-152, HarD-Net- 138s achieves the same accuracy with a GPU inference time reduction by 35%. Evaluating with ResNet-50, HarD-Net- 68 accomplishes an inference time reduction by 30%, which is also a suitable backbone model for object detections that enhances the accuracy of SSD to be higher than using ResNet-101 while the inference time is also significantly reduced from SSD-VGG.

In brief, accuracy-over-model-size along with accuracy-over-MACs tradeoffs, accuracy-over-DRAM-traffic for feature-maps is indeed an important factor when designing neural network architectures.

### D. Deep Dual Resolution Network (DDR-Net)

It is a network with deep high-resolution interpretation ability for real-time semantic segmentation of high-resolution images, focused for road-driving images. It starts with one trunk and then is divided into two parallel deep branches with different determination. One deep branch deploys relatively high-resolution feature maps and the other extracts rich contextual material by multiple down sampling operations. Multiple bilateral connections are bridged between two branches for efficient information fusion. DDR-Net attains an excellent balance between segmentation accuracy and inference speed and takes up less GPU memory than HR-Net during training.

When incorporating it with low-resolution feature maps, it leads to low increase in inference time. By simply increasing the width and depth of network, DDR-Net accomplishes a top trade-off between m-IoU and FPS among existing methods, from 77.4% m-IoU at 109 FPS to 80.4% mIoU at 23 FPS on Cityscapes test set. DDR-Net only utilizes basic residual modules and bottleneck modules and can provide a wide range of speed and accuracy trade-off by scaling model width and depth. Due to the simplicity and efficiency, it can be seen as a strong baseline towards unifying real-time and high-accuracy semantic segmentation.

## IX. DESIGN OF UNET-BASED ARCHITECTURES

Asymmetric UNet baseline and lightweight encoders are the main components of most UNet-based architectures. The UNet-based model consists of lightweight encoder such as MobileNetV2, MobileNetV3, and Efficient Net and the decoder similar to the original UNet. The encoder in UNet is called contracting path, capturing context features from an input image. The contracting path usually loses spatial information due to the gradual down sampling of the feature maps.

On the other hand, the decoder in UNet or expansion path localizes features and rebuilds the segmentation map. Concatenation blocks are supplemented to this structure, combining low-level and high-level features and improving the network's overall performance. Long skip connections pass fine-grained details from encoder to decoder to further optimize the architecture. In summary, the main components of UNet baseline architecture are lightweight encoder blocks for down sampling, decoder blocks for up sampling, concatenation blocks that fuse low-level and high-level features, and skip connections which improve accuracy while reducing computations.

## A. MobileNetV2 Encoder

MobileNetV2 utilizes depth wise divisible convolutions and the inverted residual blocks to reduce computations. Depth wise separable convolution is an optimized technique exploited in all versions of Mobile Nets. It consists of two distinct layers: depth- wise convolution and pointwise convolution. Depth-wise convolution applies a filter per channel input, and pointwise convolution creates new features by linearly combining input channels. Depth wise separable convolutions generally decrease computations compared to standard convolutional layers. Similarly, inverted residual blocks or Mobile Bottleneck Convolution (MBConv) generate fewer parameters and enhance efficiency. In this approach, the skip connections connect two narrow feature channels, skipping wider channels. Consequently, Mobile Bottleneck Convolution creates lower feature dimensions while carrying information from the earlier layers. The Inverted Residual Block is the central element in lightweight networks such as MobileNetV2, MobileNetV3, and Efficient. We develop a UNet-based pipeline made up of MobileNetV2 pretrained on ImageNet and the decoder identical to UNet [34].

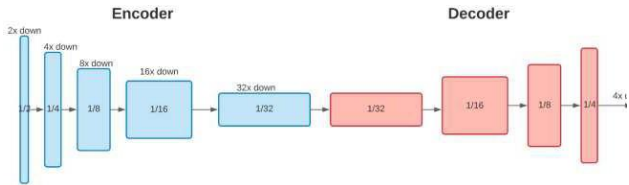

Fig. 6. Encoder Decoder Architecture.

We train the network on the Flood Net dataset and name the trained model UMBV2. The results of aerial segmentation using UMBV2 are available in the subsequent sections.

## B. MobileNetV3 Encoder

MobileNetV3 is the up- graded version of MobileNetV2. It keeps the main building blocks of MobileNetV2 and adds squeeze and excitation blocks to improve the accuracy of the network without adding extra computational costs. A squeeze-and-excitation network is a convolutional block that modifies each feature map's weights according to its significance.

In addition to the squeeze-and- excitation block, MobileNetV3 redesign some non-efficient components to reduce computations. For example, they replace the sigmoid activation function with hard sigmoid to build the most effective models. We construct a UNet-based structure with MobileNetV3 small pretrained on ImageNet. Then, we train the whole model on the Flood Net dataset and call the trained model UMBV3 [41].

## C. EFFICIENT-NET ENCODER

Efficient-Net is a family of lightweight encoders which we apply into UNet baseline architecture. Similar to Mobile Nets, the primary building blocks of Efficient-Net are bottleneck convolutions (MBConvs). In addition, squeeze-and-excitation optimization is used in the structure of Efficient-Net, similar to MobileNetV3.

The main difference to the previous methods is applying a compound scaling method that simultaneously scales each convolutional block in three dimensions of width, depth, and resolution. Efficient-Net allows scale up and scale down the network based on available computing resources. For example, the EFFICIENT-NETB0 has the minimum number of MBCONVS, and Unet has two paths:

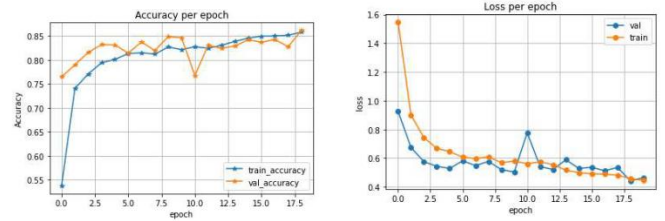

Fig. 7. Accuracy per epoch and Loss per epoch.

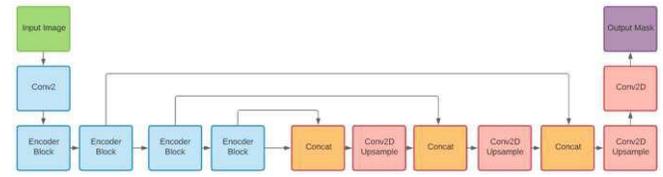

Fig. 8. Overall Structure of UNet-Baseline Model.

- The first path is contraction path also called as encoder It consists of the repeated application of two 3x3 convolutions. Each conv is followed by a Re-LU and batch normalization. Then a 2x2 max pooling operation is applied to reduce the spatial dimensions. Again, at each down sampling step, we double the number of feature channels, while we cut in half the spatial dimensions [34].

- The second path is expansion path also called as decoder. Every step in the expansive path consists of an up sampling of the feature map followed by a 2x2 transpose convolution, which halves the number of feature channels [34].

## X. SETTING UP THE JETSON XAVIER AGX

We use a Linux host computer to run Jetpack installer and flash the developer kit. We connect the developer kit and Linux host computer to the same network. The developer kit was in Force USB Recovery Mode (RCM) so the installer can transfer system software to the Jetson AGX Xavier module. Jetpack installer includes a Component Manager, allowing us to choose what to install on the Linux host computer and/or the Jetson Developer Kit.

To get started with the hardware, we wrote the Jetson Xavier NX Developer Kit (Jetpack SDK) onto a fresh microSD card. We Followed the instructions on the NVIDIA website to install the image. The Jetpack version at the time of writing was 4.5.1. Using Linux, we run the command, inserted the SD card, started up the Jetson, and clicked through the installation procedure. The training of the models was done on our modern desktop PC, with a GPU. Since the Jetson developer kit is not designed for such task and would take way too long to train even a simple model. Our host machine was running Ubuntu 18.04 / Windows 10 and had Nvidia GPU with CUDA 10.0 and cu-DNN 7.6.5 installed.

We trained our models with Py-Torch and afterwards exported to the AGX Xavier. The resulting models were converted to Tensor-RT engines. During network design, we paid attention to only use operations that are supported and highly optimized by Tensor-RT. This enables faster inference compared to pure Py-Torch. Due to the fast runtime and robust semantic segmentation, our UNET based model is well suited as a common initial processing step in a complex system for real-time scene analysis.

XI. INSTALLING PY-TORCH LIBRARIES AND DEPENDENCIES

We have installed several libraries and dependencies to get PyTorch to run.
- Numpy
- Pandas
- Scipy
- Scikit-Image
- Matplotlib
- Seaborn
- Opencv-Python
- Torch
- Torch-vision

We save the trained model on the system as python dictionary in ". pth" file Port the ". pth" file to the board using Moba-Xterm. We converted pytorch file to Tensor-RT engine compatible with on Xavier AGX board. Getting these libraries to work on an NVIDIA Jetson was not straight forward, since pre-compiled python wheels were not always available for aarch64. Pip installs triggered a compilation of the libraries, which, in the case of computer vision libraries, often requires lots of dependencies that was tricky to build correctly on ARM architectures [14]. Some libraries were pre-installed or had pre-built wheels:
- OpenCV-python: 4.1.1 is pre-installed for Python 3.6 on Jetpack 4.5.1
- Torch: 1.8 can be downloaded as a wheel from the NVIDIA forums.

Some are pre-installed but outdated and will require updating:

- Numpy: 1.13 (current 1.19.5)
- Matplotlib: (latest usable with Python 3.6 is 3.3.4)
- Pandas: 0.22.0 (current 1.1.5)
- Scipy: 0.19.1 (current 1.5.4)

XII. EVALUATION & RESULTS

We implement two semantic segmentation methods, UNet-MobileNetV2, and UNet-MobileNetV3, and evaluated their performance on the Flood-Net dataset. For UNet based architectures (UNet-MobileNetV2 and UNet-MobileNetV3), we adopt the "Adam-W" optimizer with a base learning rate of 0.001, weight decay of 0.0001. Furthermore, we used Cross-Entropy Loss as the loss function. We focus on the performance and report the intersection of union per class (IoU) and overall mean intersection over union (m-IoU) on our test set. We can consider UNet-MobileNetV2 and UNet-MobileNetV3 models as feasible real-time segmentation methods for aerial applications.

*A. Measure of Accuracy*

Intersection over Union is an evaluation metric used to measure the accuracy of an object detector on a particular dataset. In the numerator we compute the area of overlap between the predicted bounding box and the ground-truth bounding box. The IoU metric measures the number of pixels common between the target and prediction masks divided by the total number of pixels present across both masks.

The Intersection over Union (IoU) metric, also referred to as the Jaccard index, is essentially a method to quantify the percent overlap between the target mask and our prediction output. An alternative metric to evaluate a semantic segmentation is to simply report the percent of pixels in the image which were correctly classified. The pixel accuracy is commonly reported for each class separately as well as globally across all classes.

When considering the per-class pixel accuracy we are essentially evaluating a binary mask; a true positive represents a pixel that is correctly predicted to belong to the given class (according to the target mask) whereas a true negative represents a pixel that is correctly identified as not belonging to the given class [34].

The intersection of union (IoU) measures the similarity of predicted masks to the ground truth. We measure IoU at pixel-level for nine classes including Non-flooded Building (NB), Flooded Building (FB), Non-flooded Road (NR), Flooded Road (FR), Water, Tree, Vehicle, Pool, and Grass by using the following formula:

$$IoU = \frac{TP}{TP+FP+FN} \qquad (1)$$

The mean Intersection assesses semantic segmentation methods over union (m-IoU) on the Flood Net dataset [15]. We used mean intersection over union (m-IoU) for all nine classes by applying the following formula:

$$mIoU = \frac{1}{K}\sum_{i=0}^{k} \frac{TPi}{TPi+FPi+FNi} \qquad (2)$$

TABLE IV. PERFORMANCE IN TERMS OF ACCURACY (MIOU). IOU OF CLASSES INCLUDING NON-FLOODED BUILDING (BN), FLOODED BUILDING (BF), NON-FLOODED ROAD (NR), FLOODED ROAD (FR), WATER, TREE, VEHICLE, POOL, AND GRASS.

| Model | Image Size | BF | BN | FR | NR | Water | Tree | Vehicle | Pool | Grass |
|---|---|---|---|---|---|---|---|---|---|---|
| UNet-MobileNetV2 | 512x512 | 43.5 | 59.3 | 21.2 | 61.2 | 73.3 | 64.9 | 15.1 | 32.7 | 82.8 |

## B. Measure of Efficiency

Computational cost is simply a measure of the amount of resources the neural network uses in training or inference, which is important so you can know how much time or computing power needed to train or use an NN. It can measure in a variety of ways, but common ones are time and number of computations, expressed either as number of floating-point operations (FLOPs) or as number of multiply-and-accumulate operations (MACs or MACCs) the multiply-accumulate operation is a common step that computes the product of two numbers and adds that product to an accumulator. The hardware unit that performs the operation is known as a multiplier–accumulator (MAC, or MAC unit); the operation itself is also often called a MAC or a MAC operation [14].

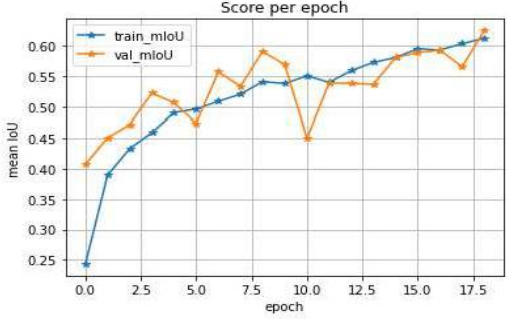

Fig. 9. Score per epoch.

Basically, the architecture of MAC is classified as parallel, recursive, and shared segmented structure. The multiply-accumulation is achieved using iterative calculation of smaller module through several clock cycles. The latency and throughput of the MAC depends on the number of multipliers and adders, which are recursively called for each cycle [34].

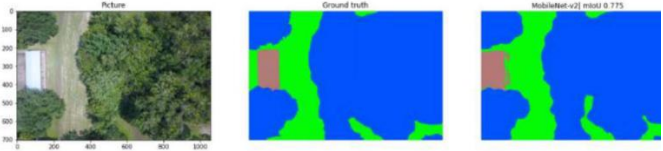

Fig. 10. Visual comparison of real-time segmentation using Mobile-NetV2 on Flood Net test dataset1.

In the first cycle, data is fetched from the internal memory, the second cycle involves multiplication process, during the third cycle summation takes place, while in the fourth cycle the function of A.B+Acc is performed and finally the last output is latched within the internal memory. This approach utilizes minimum hardware by using reusability of resources with increased latency. These types of MAC architectures are deployed in embedded Advanced RISC (Reduced-Instruction-Set-Computing) Machine (ARM) core [25].

TABLE V. PERFORMANCE IN TERMS OF EFFICIENCY

| Deep Network | Parameters | MAC |
|---|---|---|
| UNet+ MobileNetV2 | 6.63M | 13.87G |

The trained networks were converted to Tensor-RT engines to activate jetson nano GPUs and for optimizing the runtime. All latency and power measurements include only model inference time for iterating through 1000 static frames. The measurements only refer to latency and power for frame processing. All measurements were collected for 1000 sequential frames over 20 rounds for generalization [34].

TABLE VI. PERFORMANCE COMPARISON BETWEEN CPU/GPU

| UN-ET based Mobile-Net-V2 (Video 915 frames), Resolution 256 X 256 (MIOU: 61.9%) | | | | | | | | |
|---|---|---|---|---|---|---|---|---|
| Device Specs | Param (Millions) | Model Size (MB) | Comps (GOPS) | Latency (s) | Thp (FPS) | Avg Pwr (W) | FPS/W | GOP/J |
| Nano-CPU | 6.6 | 26.4 | 3.45 | 6.19 | 0.16 | 2.3 | 0.07 | 0.48 |
| Nano-GPU | | | | 0.07 | 15.12 | 5.9 | 2.56 | 16.71 |

## XIII. ISSUES AND CHALLENGES

Our project shows that there is still a gap between real-time and standard semantic segmentation models with a dilation backbone.

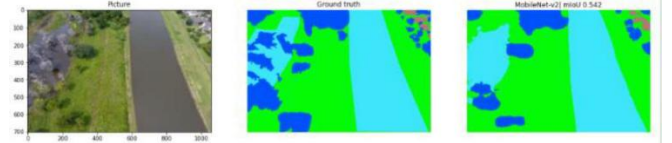

Fig. 11. Visual comparison of real-time segmentation using Mobile-NetV2 on Flood Net test dataset 2.

However, we believe that real-time architectures have the potential to improve their accuracy by scaling up encoders. One of the main findings of our experiment is that we can increase the network's performance by upgrading the encoder, as we boost the network by replacing MobileNetv2 with MobileNetV3. We can apply these findings to our future work.

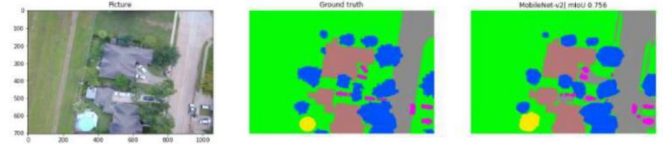

Fig. 12. Visual comparison of real-time segmentation using Mobile-NetV2 on Flood Net test dataset 3.

One of the crucial aspects of our semantic segmentation models are the ability to segment damaged areas. As expected, all models demonstrate lower performance for the segmentation of flooded buildings and flooded roads. Above all, PSP-Net segments the flooded buildings and roads with the test IoU of 68.9% and 82.2%, respectively.

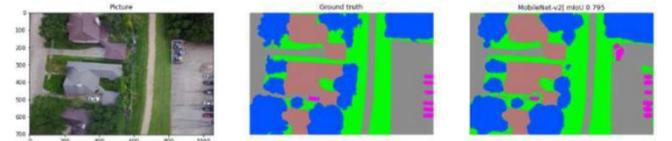

Fig. 13. Visual comparison of real-time segmentation using Mobile-NetV2 on Flood Net test dataset 4.

UNet-Mobile-NetV2 yields 58.5% IoU for flooded buildings and UNetMobileNetV3 attains 43.6% IoU for flooded roads. It is a significant result because real-time segmentation methods will provide a way to classify damaged areas in catastrophic events [34].

It is difficult to implement the UNet based models on edge devices like NVIDIA Jetson Xavier and we can try to experiment this with other edge devices like Coral Edge TPU etc.

XIV. Conclusion and Future Work

We successfully trained four models on the aerial imagery dataset –Flood-Net and obtained qualitative and quantitative results on all our models. This work is a benchmark for semantic segmentation of aerial imagery during natural disasters such as flooding events. Furthermore, we apply our real-time semantic segmentation models on videos, and we can see real-time segmentation in different classes. Our future work is to improve our real-time segmentation model by scaling up the encoder to enhance the performance on deformed and damaged areas [34].

This result can benefit UAV systems for real-time emergency response during disasters. In the future we can deploy the UNET based model to Unmanned Aerial Vehicle (UAV). Networks with high accuracy on everyday images such as those of ImageNet cannot really be used to detect image features from aerial datasets which contain more complex urban and natural scenes. Thus, there is a need to design separate novel architectures which can effectively detect urban disasters. More work is needed for security aspects towards deployment of the UNET based models [34].

In the future, semantic segmentation and wireless communication technologies have revolutionary potential for a myriad of applications. By deploying the latest semantic segmentation technologies and other recent machine learning achievements in UAV, paired with wireless communication technologies for coordination and communication to form a sound UAV communication architecture, and paired with IOT sensors for various protocols [45-55] for efficient parallel communication [56-68], we can build an effective, decentralized, parallel communication network [69-80].

Acknowledgment

This project was funded by Amazon, Microsoft, Bina lab, and ARL under Grant W911NF21-20076. We want to offer our gratitude to Prof. Ting Zhu for guiding us and helping us to accomplish this project work. We also want to thank to Prof. Maryam Rahnemoonfar for providing us the gudiance and resources at Computer Vision and Remote Sensing, BINA Lab at Lehigh University. We also want to thank Professor Robin Murphy at Texas A&M University for providing the raw images after Hurricane Harvey.